\documentclass[10pt,twocolumn,letterpaper]{article}

\newif\ifwithsuppl
\withsuppltrue

\newcommand{\IfWithSupplTF}[2]{\ifwithsuppl#1\else#2\fi}

\usepackage[pagenumbers]{cvpr} %
\usepackage{amsthm,amsfonts,amsmath,amssymb}
\usepackage{multirow}
\usepackage{graphicx}
\usepackage[table,xcdraw]{xcolor}

\definecolor{cvprblue}{rgb}{0.21,0.49,0.74}
\usepackage[pagebackref,breaklinks,colorlinks,allcolors=cvprblue]{hyperref}
\usepackage{lipsum}

\title{From Rays to Projections: Better Inputs for Feed-Forward View Synthesis}

\author{Zirui Wu$^1$
\quad
Zeren Jiang$^2$
\quad
Martin R. Oswald$^3$
\quad
Jie Song$^{1,4}$\\
$^{1}$HKUST (GZ)\quad
$^{2}$University of Oxford\quad
$^{3}$University of Amsterdam\quad
$^{4}$HKUST\\
\small\url{https://wuzirui.github.io/pvsm-web}
}

\makeatletter
\renewcommand\paragraph{\@startsection{paragraph}{4}{\z@}%
                                    {0.5ex\@plus 0.3ex \@minus 0.2ex}%
                                    {-1em}%
                                    {\normalfont\normalsize\bfseries}}

\makeatother

\begin{document}

\twocolumn[{%
\renewcommand\twocolumn[1][]{#1}%
\maketitle
\vspace{-3em}
\begin{center}
    \centering
    \captionsetup{type=figure}
    \includegraphics[width=\textwidth]{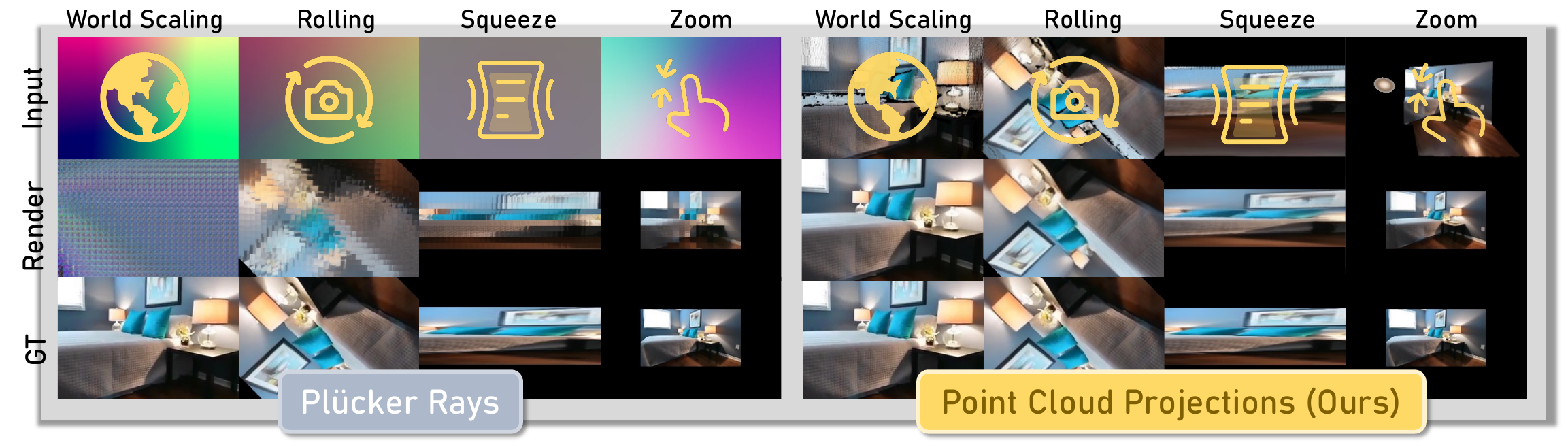}
    \captionof{figure}{\textbf{Projective conditioning enables robust novel view synthesis.}
    We investigate what camera pose encoding best conditions a view synthesis model. Compared to the commonly used absolute valued Pl\"ucker ray conditioning by prior works~\cite{jin_lvsmlarge_2025,jiang_rayzerselfsupervised_2025}, our proposed projection conditioning encodes scene-camera configuration as a relative transformation. Under various geometric transformations, this shows better robustness while the absolute conditioning signal fails due to the non-smoothness of transformations in the Pl\"ucker ray space.
    }
    \label{fig:teaser}
\end{center}%
}]

\begin{abstract}
Feed-forward view synthesis models predict a novel view in a single pass with minimal 3D inductive bias. Existing works encode cameras as Plücker ray maps, which tie predictions to the arbitrary world coordinate gauge and make them sensitive to small camera transformations, thereby undermining geometric consistency. In this paper, we ask what inputs best condition a model for robust and consistent view synthesis. We propose projective conditioning, which replaces raw camera parameters with a target-view projective cue that provides a stable 2D input. This reframes the task from a brittle geometric regression problem in ray space to a well-conditioned target-view image-to-image translation problem. Additionally, we introduce a masked autoencoding pretraining strategy tailored to this cue, enabling the use of large-scale uncalibrated data for pretraining. Our method shows improved fidelity and stronger cross-view consistency compared to ray-conditioned baselines on our view-consistency benchmark. It also achieves state-of-the-art quality on standard novel view synthesis benchmarks.
\vspace{-7mm}
\end{abstract}
    
\section{Introduction}
\label{sec:intro}

Synthesizing realistic novel views from a set of captured context images is a long-standing goal in computer vision and graphics. Recent feed-forward models~\cite{yu_pixelnerfneural_2021,charatan_pixelsplat3d_2024,smart_splatt3rzeroshot_2024,jin_lvsmlarge_2025} leverage data-driven priors to directly render novel views in a single forward pass, by conditioning the model on a few context views and the target camera frustum.

In particular, the recent large view-synthesis models (LVSMs)~\cite{jin_lvsmlarge_2025,jiang_rayzerselfsupervised_2025,wang_lessyou_2025} employ vision transformers (ViTs)~\cite{dosovitskiy_imageworth_2021} and encode camera parameters using Pl\"ucker ray embeddings~\cite{plucker_xviinew_1997} as their input space representation. While this interface is convenient, it can also introduce brittleness. The absolute coordinate representation makes the model sensitive, such that even minor adjustments to the camera can lead to significant shifts in the 6D ray space, despite only small visual changes. This sensitivity can lead to inputs straying from the model's training distribution, which in turn degrades 3D consistency, as illustrated in \cref{fig:teaser}. Basic transformations, such as zooming and squeezing, can be combined to create complex camera movements. For example, simultaneously adjusting the zoom while moving the camera produces the dolly zoom effect (see Suppl. video). However, existing models struggle to robustly handle these fundamental operations, resulting in artifacts that violate geometric consistency.

In this work, we investigate the input representations for feed-forward novel view synthesis and pose the question: \emph{What inputs best condition a model for robust and consistent view synthesis?} Instead of directly encoding camera parameters, we propose \textbf{projective conditioning}, an approach that involves supplying the model a \emph{point cloud projection image}.  This image is generated by first extracting depth maps from context views using off-the-shelf perception models ~\cite{keetha_mapanythinguniversal_2025} and then rasterizing the unprojected point cloud into the target camera view. This method delegates camera handling to a deterministic geometry engine, allowing the model to operate within a stable 2D image domain. 

The advantages of this representation are twofold:
\textit{First}, minor adjustments to the camera lead to correspondingly small and localized changes in the point cloud image, facilitating more precise camera control and enhancing robustness to various camera transformations. This approach reframes novel view synthesis as an image-to-image mapping conditioned on coherent visual cues, ultimately resulting in stronger 3D consistency across different viewpoints.
\textit{Second}, since projective conditioning operates on 2D buffers, it can be seamlessly combined with Masked Auto-Encoding (MAE)-style pretraining~\cite{he_maskedautoencoders_2021,weinzaepfel_crocoselfsupervised_2023}. We leverage the observation that the task of reconstructing a randomly masked target image structurally mirrors our view synthesis goal and propose a self-supervised pretraining strategy, which allows the model to learn a powerful cross-view completion prior.

To evaluate rendering consistency, we establish a view consistency benchmark using common camera transformations in \cref{fig:teaser}, applied to the target view. Our experiments reveal that, with identical backbones and training schedules, projective conditioning exhibits enhanced fidelity and stronger cross-view consistency compared to ray-based conditioning. Thanks to the rapid advancements in recent geometric foundation models~\cite{wang_dust3rgeometric_2023,wang_vggtvisual_2025,keetha_mapanythinguniversal_2025,liu_worldmirroruniversal_2025}, we can now extract high-quality 3D annotations from uncalibrated images. To facilitate training and evaluation at scale, we have curated a derivative of the RealEstate10K~\cite{re10k} dataset, which includes dense depth information and refined camera parameters using MapAnything~\cite{keetha_mapanythinguniversal_2025}.

Our contributions can be summarized as follows:
\begin{itemize}
\item We analyze the instability of Pl\"ucker ray conditioning and propose to use projective conditioning as a stable 2D input representation. This approach reframes view synthesis as an image-to-image mapping, enhancing camera control and robustness to various transformations.
Extensive experiments demonstrate that our model outperforms existing ray-conditioned baselines in both view consistency and novel view synthesis quality. Code, data, and models will be made publicly available
\item We introduce a MAE pretraining stage that leverages the 2D nature of projective conditioning to learn scene completion priors in a self-supervised manner, thereby reducing dependence on expensive 3D annotations.
\item We contribute a refined RealEstate10K dataset, which includes dense 3D annotations, along with a view consistency benchmark for rigorous evaluation of robustness.
\end{itemize}

\begin{figure*}[t]
    \centering
    \includegraphics[width=\linewidth]{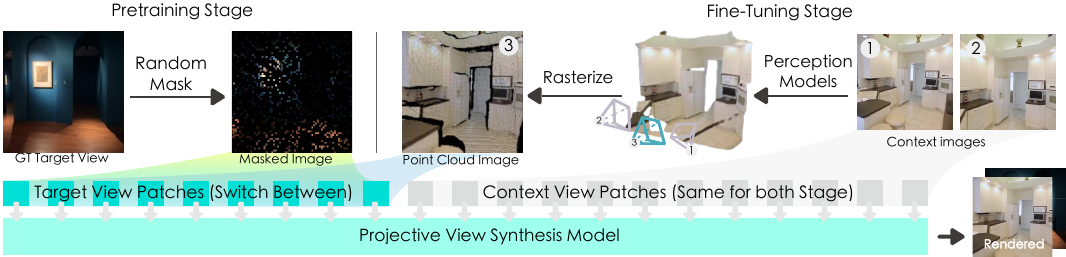}
    \caption{\textbf{An overview of our proposed two-stage training pipeline}. 1. \textbf{Pretraining}: This stage is self-supervised with the model conditioned on a set of context views and a randomly masked version of the target view itself (Masked Image). Its objective is to reconstruct the complete, original Ground Truth (GT) Target View. 2. \textbf{Fine-Tuning}: The context views are first unprojected into a unified 3D point cloud with extracted depth from perception models~\cite{keetha_mapanythinguniversal_2025}, which is then rasterized from the perspective of the target camera's frustum to create a point cloud projection image that provides geometric cues. The model is then fine-tuned to generate the final target image.}
    \label{fig:pipeline}
    \vspace{-3mm}
\end{figure*}

\section{Related Works}
\label{sec:related-works}

\paragraph{Neural Rendering.} 
Novel view synthesis has long been a central task in 3D computer vision. Neural representations such as neural radiance fields (NeRFs)~\cite{mildenhall_nerfrepresenting_2020,barron_zipnerfantialiased_2023, ye2024nerf, liu2024ripnerfsiggraph} and 3D Gaussian Splattings~\cite{bernhard_3dgaussian_2023,ye_gaustudiomodular_2024,wu_3dgaussian_2025, ye2025gs,yang2024spectrally,lou2025roboicra,zheng2024gaussiangrasper} have achieved impressive results with volumetric scene parameterizations. Their main limitation is the need for per-scene optimization, which leads to high computation cost and weak generalization beyond the training scene.

\paragraph{Feed-forward View Synthesis.}
To alleviate this cost, feed-forward alternatives have gained traction.
pixelNeRF~\cite{yu_pixelnerfneural_2021} conditions NeRFs on context views without to predict novel views in a single forward pass instead of optimizing a scene-specific model.
Follow-up work on feed-forward 3D Gaussians~\cite{charatan_pixelsplat3d_2024,chen_mvsplatefficient_2024,ye_nopose_2024,jiang_anysplatfeedforward_2025,liu_worldmirroruniversal_2025} regresses pixel-aligned Gaussian primitives from context views and splats them into the target camera.
More recent methods move toward implicit networks that avoid explicit 3D structures. 
LVSM~\cite{jin_lvsmlarge_2025} and its extensions~\cite{jiang_rayzerselfsupervised_2025,wang_lessyou_2025,li_camerasrelative_2025} directly map camera ray embeddings to target RGBs, demonstrating strong scalability.
Our method instead operates entirely in the 2D image domain, predicting the target view from a projective point-cloud cue rendered in the target camera, without explicit pose inputs or assumptions about an underlying 3D representation.
Apart from our method, recent works~\cite{yu_viewcraftertaming_2025,ren2025gen3c,gu2025das, yan2024streetcrafter,yu_trajectorycrafterredirecting_2025} also work with projective cues, but mainly use them to drive 3D-aware video generation models that stochastically complete unseen regions via iterative denoising, whereas we use a deterministic single-pass regressor for efficient novel view synthesis.

\paragraph{Self-supervised Learning.}
Self-supervised learning (SSL)~\cite{he_maskedautoencoders_2021,dinov3,2022JPRS..194..302T,weinzaepfel_crocoselfsupervised_2023} is an appealing way to exploit large-scale unlabeled data.
MAE~\cite{he_maskedautoencoders_2021} masks a large portion of image patches and trains the model to reconstruct them, yielding 2D priors that transfer well to downstream tasks.
In 3D vision, where dense geometry is often scarce or noisy, SSL is especially valuable.
CroCo~\cite{weinzaepfel_crocoselfsupervised_2023} extends masked modeling to multi-view images, and RayZer~\cite{jiang_rayzerselfsupervised_2025} and Less3D~\cite{wang_lessyou_2025} apply SSL to feed-forward view synthesis by mapping uncalibrated images to a latent SE(3) manifold and predicting target views from context images and latent cameras.
This reframes novel view synthesis as an SSL problem and scales well, but also introduces two limitations: (1) the model can indirectly access the ground-truth target view through the latent-camera construction, which risks information leakage; and (2) the latent manifold is not aligned with a physical coordinate system, making precise viewpoint control difficult in practice.
In contrast, we treat the projected point-cloud cue as an explicit structural signal, pre-training with a MAE-style objective on DL3DV~\cite{ling2024dl3dv} and then fine-tuning the network to map this cue directly to the target RGB view.

\section{Analysis of Ray Conditioning Instability}
\label{sec:input-analysis}
We begin by reviewing the core pipeline of the Large View Synthesis Model (LVSM)~\cite{jin_lvsmlarge_2025}, which serves as the foundation for our analysis. LVSM formulates novel view synthesis as a single-pass, conditional image regression task using a Vision Transformer (ViT) architecture~\cite{dosovitskiy_imageworth_2021}. 
Given $N^c$ context views $\{\mathcal{I}^c_i\}^{N^c}_{i=1}$ with known camera parameters $\{\mathcal{C}^c_i\}^{N^c}_{i=1}$, and a target camera pose $\mathcal{C}^t$, the model predicts the target image $\mathcal{I}^t$ via a mapping $\mathcal{\hat{I}}^t = \mathcal{M}(\{(\mathcal{I}^c_i, \mathcal{C}^c_i)\}^{N^c}_{i=1}, \mathcal{C}^t)$. Central to this approach is its use of Pl\"ucker coordinates.

\subsection{Preliminaries: Large View Synthesis Models}
\label{sec:lvsm-pipeline}

\paragraph{Pl\"ucker Ray Representation.}
LVSM first converts each camera pose $\mathcal{C} = (K, [R|t])$ (including the intrinsics $K$ and extrinsics $[R|t]$) into a per-pixel 6D Pl\"ucker coordinate. The viewing ray of a pixel is defined by the ray origin $\mathbf{o}$ and normalized direction vector $\mathbf{d}$. The Pl\"ucker representation is then constructed as $\mathbf{L} = (\mathbf{m}, \mathbf{d})$, where $\mathbf{m} = \mathbf{o} \times \mathbf{d}$ is the moment vector.
All valid Pl\"ucker coordinates form the Klein quadric $\mathcal{K}$, which is the embedding of the Grassmannian manifold $G(1, 3)$ into $\mathbb{P}^5$~\cite{griffiths2014principles}.

\paragraph{Pipeline.}
Both context images and their associated Pl\"ucker maps are split into non-overlapping $p \times p$ patches and linearly embedded into tokens. The target view ray map is also split into patches and linearly embedded into tokens similarly.
All tokens are concatenated and processed by a decoder-only Vision Transformer~\cite{vaswani_attentionall_2017,dosovitskiy_imageworth_2021}, whose output tokens corresponding to the target view are decoded into RGB patches.
While effective in practice, this approach encodes all geometric structure through absolute Pl\"ucker coordinates. As shown below, this induces instability under coordinate-gauge changes.

\subsection{The Global SE(3) Invariance}
\label{sec:se3-invariance}
A fundamental property any view-synthesis model should satisfy is the invariance to the choice of world coordinate gauge.
Formally, the rendered image $\mathcal{I}^t$ should remain unchanged if the entire scene geometry $\mathcal{G}$ and all cameras $\{\mathcal{C}_i\}$ are all transformed by a global transformation $g \in \mathrm{SE}(3)$:
\begin{equation}
    \label{eq:se3-invariance}
    \mathcal{M}(\{(\mathcal{I}^c_i, g\cdot \mathcal{C}^c_i)\}^{N^c}_{i=1}, g\cdot\mathcal{C}^t) =\mathcal{M}(\{(\mathcal{I}^c_i, \mathcal{C}^c_i)\}^{N^c}_{i=1}, \mathcal{C}^t)
\end{equation}

When expressed in Pl\"ucker space, the action $\rho(g)$ corresponding to the transformation $g=(R,\mathbf{t})$ applies as:
\begin{equation}
\label{eq:se3-invariance-action}
(\mathbf{m}', \mathbf{d}')=\rho(g)(\mathbf{m}, \mathbf{d})=(R \mathbf{m}+[\mathbf{t}]_\times R \mathbf{d},\, R \mathbf{d}).
\end{equation}

This shows that even a uniform world-space transformation can cause non-uniform, spatially varying perturbations in Pl\"ucker coordinates.
Each pixel-level ray token changes differently depending on its location, making the representation highly sensitive to the arbitrary chosen world gauge. In practice, identical relative camera-scene configurations can yield drastically different Pl\"ucker distributions if expressed in different coordinate frames.

\begin{figure}[t]
    \centering
    \includegraphics[width=\linewidth]{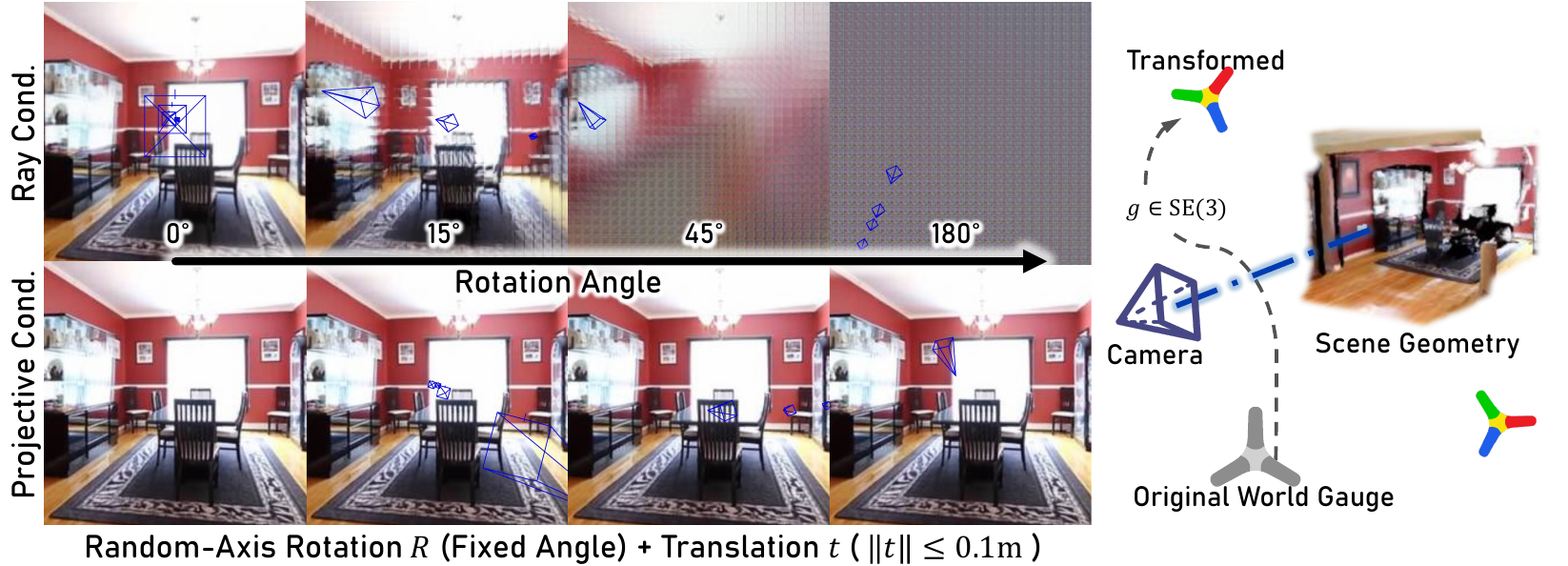}
    \caption{Under a random global $\mathrm{SE}(3)$ transformation to the global coordinate system, ray-conditioned models~\cite{jin_lvsmlarge_2025} produce degenerate results while projective conditioning remain robust.}
    \label{fig:random-coordinates}
    \vspace{-5mm}
\end{figure}

As visualized in \cref{fig:random-coordinates}, a small random $\mathrm{SE}(3)$ perturbation can cause severe degradation in Pl\"ucker-conditioned models.
This reveals that the network tends to overfit to dataset-specific coordinate gauges, leading to a noticeable train–test gap. Moreover, the absolute world reference frame carries no meaningful information for the rendering problem, so learning the invariance (e.g., by data augmentation) only wastes model capacity and training resources.

\subsection{Fine-grained Invariance}
\label{sec:generalized-invariance}
Beyond global rigid motions, rendering should remain invariant to broader transformations that preserve the underlying ray–scene relationships.
For a given scene $\mathcal{G}$ and ray $\mathbf{r}$, the pixel color determined by their intersection should be unaffected by how the world coordinates, focals, or image lattice are parameterized.

\begin{figure}[t]
    \centering
    \includegraphics[width=1\linewidth]{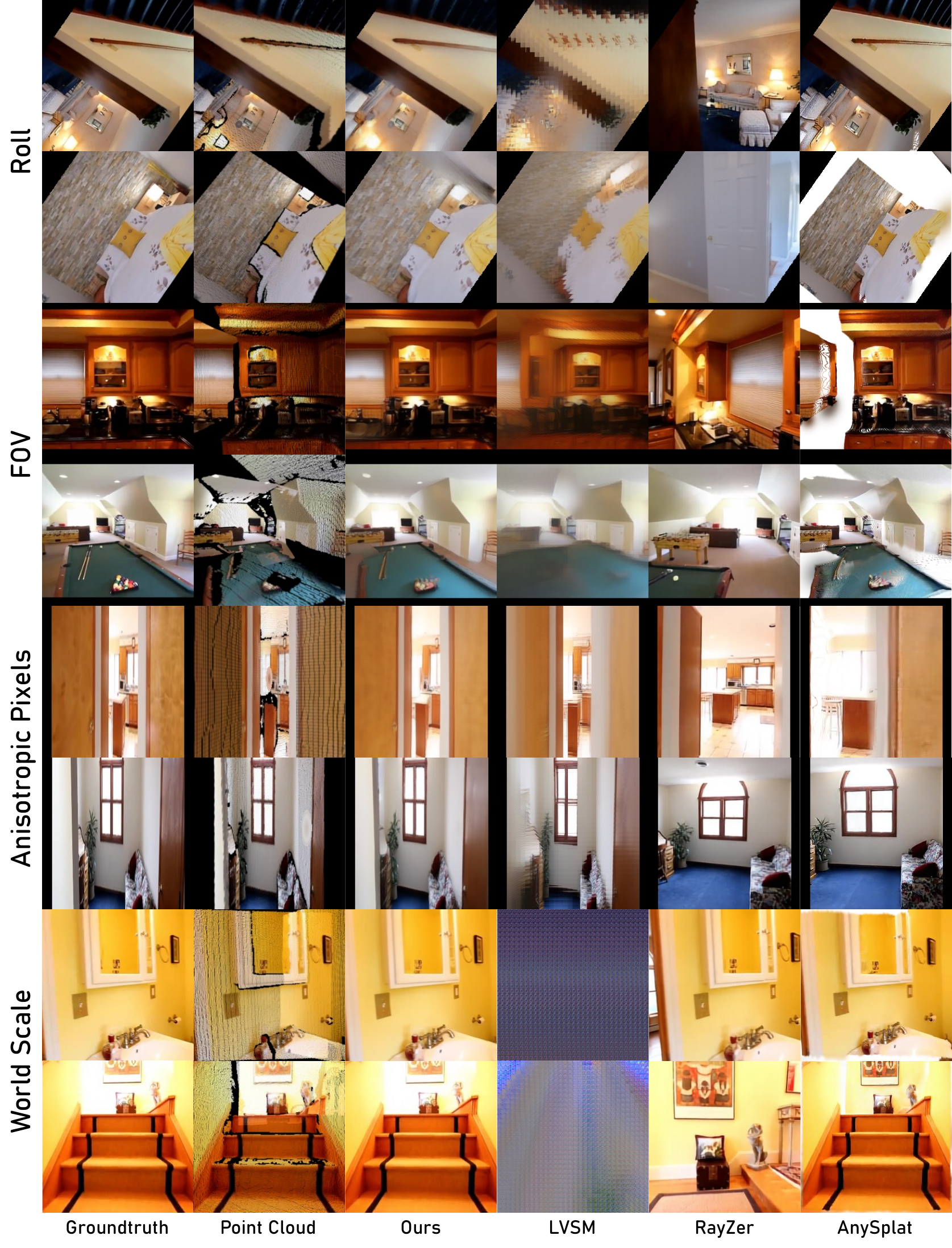}
    \caption{\textbf{Qualitative Results on our Consistency Benchmark.} Our method produces more geometrically consistent results. LVSM struggles to maintain geometric consistency, while RayZer and AnySplat fail to retrieve accurate camera parameters.}
    \label{fig:qualitative-benchmark}
    \vspace{-5mm}
\end{figure}

To examine this property, we extend the analysis from global $\mathrm{SE}(3)$ motions to include world rescaling, focal-length variation, and image-plane rotation, which are common instances of $\mathrm{Sim}(3)$ and image-space resampling.
Analogous to Eq.~\eqref{eq:se3-invariance-action}, these transformations act non-uniformly in Pl\"ucker space, yielding heterogeneous and non-local perturbations across tokens.
Empirically, these perturbations translate to large performance drops as shown in \cref{fig:qualitative-benchmark}.
This further reinforces that Pl\"ucker rays encode an over-parameterized, gauge-dependent representation ill-suited for consistent rendering.

\section{Projective View Synthesis Model}
\label{sec:method}

To circumvent the instabilities of ray conditioning, we propose the Projective View Synthesis Model (PVSM, \cref{fig:pipeline}). Our key idea is to delegate the complex geometric transformations to a deterministic rasterization engine, thereby reframing the problem from a challenging geometric regression task to a consistent image-to-image translation task.

\paragraph{Projective Conditioning.} We augment the context views with their corresponding depth maps $\{\mathcal{D}_i^c\}$, which can be readily obtained from off-the-shelf models~\cite{wang_vggtvisual_2025,keetha_mapanythinguniversal_2025}. We then warp the context images into the target view by first unprojecting their pixels into a unified 3D point cloud and then rasterizing this point cloud from the target view:
\begin{equation}
    \mathcal{I}^{c\rightarrow t}=\mathtt{Rast}(\{\mathtt{UnProj}(\mathcal{D}_i^c,\mathcal{I}_i^c,\mathcal{C}_i^c)\}, \mathcal{C}^t),
\end{equation}
where $\mathtt{UnProj}$ is the standard un-projection operator, and $\mathtt{Rast}$ is a point cloud rasterizer\footnote{We use the $\mathtt{gsplat}$~\cite{ye_gsplatopensource_2024} rasterizer due to its excellent compatibility with PyTorch tensors and batchified rendering APIs. Each 3D point is rendered as a 3D Gaussian splat~\cite{bernhard_3dgaussian_2023} with preset, fixed parameters.}. 

This reframes novel view synthesis as an image-to-image task. The resulting point cloud projection image $\mathcal{I}^{c\rightarrow t}$ (\cref{fig:pipeline} Top-right) provides a direct and coherent visual cue about the scene's geometry from the target viewpoint, explicitly handling occlusions and disocclusions. Consequently, small changes in the target camera pose $\mathcal{C}^t$ lead to smooth and localized changes in the input projection image. This grants the model inherent robustness to various camera transformations (e.g., changes in focal length, aspect ratio, or extrapolated extrinsics, etc.).

\paragraph{Architecture.}
Following previous works~\cite{jin_lvsmlarge_2025}, our model employs a decoder-only ViT as its backbone. The input token sequence is constructed from three sources: 1) patch-wise tokens from the context images ${\mathcal{I}^c_i}$, 2) patch-wise tokens from the generated point cloud projection image $\mathcal{I}^{c\rightarrow t}$, and 3) features from the pretrained DINOv3 model~\cite{dinov3} $\mathbf{f}^\mathtt{dino}$ of the context views, which we find empirically enhances structural consistency (\cref{sec:ablations}).

Formally, we tokenize each context and target view patches separately with linear embedding layers:
\begin{equation}
\mathbf{x}^\mathtt{c}_{ij} = \mathtt{Linear}_\mathtt{c}(\mathcal{I}_{ij}^c) \quad \mathbf{x}^\mathtt{t}_{j} = \mathtt{Linear}_\mathtt{p}(\mathcal{I}^{c\rightarrow t}_j),
\end{equation}
where $\mathcal{I}_{j}$ denotes the $j$-th patch of $\mathcal{I}$.
All tokens are then concatenated sequentially into $\{\mathbf{z}_i^0\} = [\{\mathbf{x}^\mathtt{c}_{ij}\}, \{\mathbf{f}^\mathtt{dino}_{ij}\}, \{\mathbf{x}^\mathtt{t}_{j}\}]$ and processed through:
\begin{align}
    \{\mathbf{z}_i^l\} &= \mathtt{TransformerLayer}^l(\{\mathbf{z}_i^{l-1}\}).
\end{align}
The output tokens from the last layer that correspond to the target view are decoded into RGB patches using a linear layer followed by a sigmoid activation, resulting in the final rendered image $\mathcal{\hat I}_j^t = \sigma\left(\mathtt{Linear}_\mathtt{o}(\mathbf{z}_j^L)\right)$.

A unique challenge with projective conditioning is that the image $\mathcal{I}^{c\rightarrow t}$ often contains empty regions. After patchification, these empty patches can lead to identical input tokens, creating ambiguity for the permutation-invariant self-attention mechanism. To resolve this issue, we apply Rotary Positional Embeddings (RoPE)~\cite{su_roformerenhanced_2023} to all tokens, ensuring that each token is assigned unique positional information.

\subsection{Quotient-Space Interpretation}
\label{sec:quotient-space}
Here we provide a quotient-space interpretation of why projective conditioning guarantees the invariance as described in \cref{sec:se3-invariance}.
Let $q = \mathtt{Rast} \circ \mathtt{UnProj}$ denote the \emph{projective conditioning operator}, which converts the full configuration space $\mathcal{X}$ into a point cloud image from the target viewpoint:
\begin{equation}
q: \mathcal{X}=\left(\big\{(\mathcal{I}_i^c,\mathcal{D}_i^c,\mathcal{C}_i^c)\big\}_i \;\times\; \{\mathcal{C}^t\}\right) \to \mathrm{Im}(q).
\end{equation}

Given that for any 3D point $\mathbf{X}\in\mathbb{P}^3$ and camera projection matrix $\mathbf{P}$, the projective image relationship holds when simultaneously apply the transformation $T$ to both:
\begin{equation}
\label{eq:projective-image-relationship}
\mathbf{P}'\,\mathbf{X}' \;\sim\; (\mathbf{P}T^{-1})(T\mathbf{X}) \;=\; \mathbf{P}\mathbf{X},
\end{equation}
which means that the transformation $T$ does not change the relationship between the 3D point and the camera.
Therefore, the composition operator $q$ produces outputs that depend only on the relative arrangement among cameras and geometry. Consequently, $q(\mathcal{X})$ provides an invariant representation of the quotient space $\mathcal{X}/\mathrm{SE}(3)$, meaning that all configurations that differ only by a global coordinate gauge are map to the same element in $\mathrm{Im}(q)$.

In practice, the network learns a mapping $h: \mathrm{Im}(q) \rightarrow \mathcal{I}^t$, and the overall function can be expressed as $f = h \circ q$. This formulation ensures \emph{gauge-free conditioning by design}: the model never needs to infer invariance to global transformations from data, since such invariance is already encoded in the projective operator itself.

\begin{figure}[t]
    \centering
    \includegraphics[width=1\linewidth]{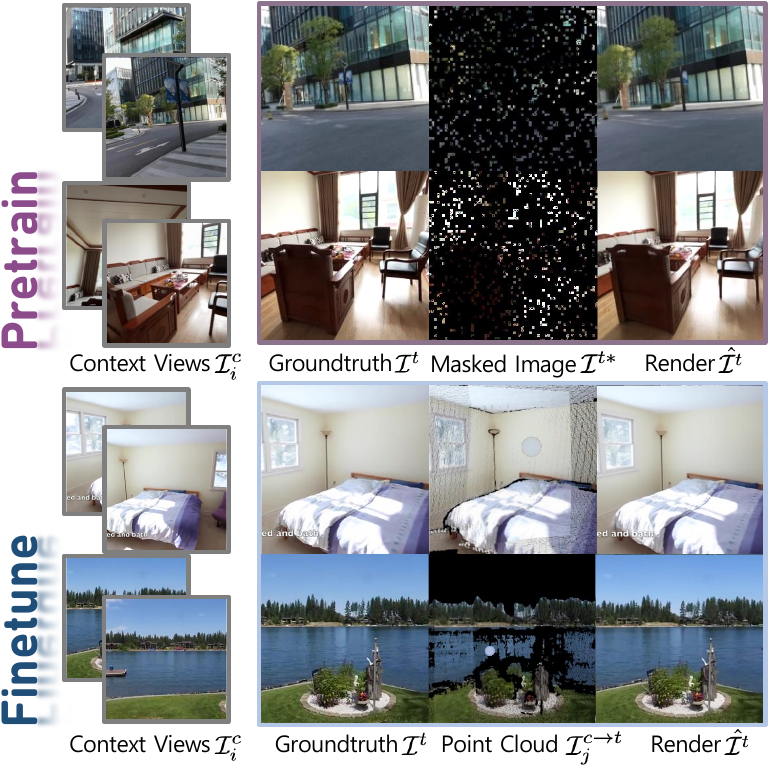} 
    \caption{\textbf{Our training stages.}
    \textit{Pretraining (top):} The model reconstructs a target view from a randomly masked version of itself, conditioned on context views, using uncalibrated image data.
    \textit{Fine-tuning (bottom):} The model is then fine-tuned to reconstruct the target view from a point-cloud projection image obtained by warping the context views into the target camera frustum.}
    \label{fig:pretrain-finetune}
    \vspace{-5mm}
\end{figure}

\subsection{Mask Auto-Encoding Pretraining}
\label{sec:mae-training}
Acquiring large-scale, calibrated RGB-D datasets for training is often resource-intensive. To mitigate this dependency and leverage abundant uncalibrated image and video data, we propose a self-supervised pretraining strategy.

Our key observation is that the projected sparse point-cloud projection image that we use visually resembles a randomly masked target image (\cref{fig:pretrain-finetune}). This motivates a pretext task inspired by masked image modeling~\cite{he_maskedautoencoders_2021,weinzaepfel_crocoselfsupervised_2023}.
During pretraining, we corrupt the ground-truth target image $\mathcal{I}^t$ to obtain $\mathcal{I}^{t*}$: first by randomly removing a portion of patches, then sparsifying some of the remaining ones by dropping pixels, and finally applying a random affine color transform to mimic exposure or camera-setting changes.

The corrupted image $\mathcal{I}^{t*}$, together with the context views ${\mathcal{I}^c_i}$, is fed into our rendering ViT, which is trained to reconstruct the original target image. This objective teaches the model robust cross-view image completion prior. After this self-supervised stage, we fine-tune the model on the projective conditioning task for a shorter schedule, yielding better performance and improved data efficiency.

\paragraph{Optimization.}
Following prior work~\cite{jin_lvsmlarge_2025, zhang_gslrmlarge_2024}, we train our model with perceptual losses:
\begin{equation}
    \mathcal{L}=\mathtt{MSE}(\mathcal{I}^t, \hat{\mathcal{I}^t}) + \lambda\cdot\mathtt{Perceptual}(\mathcal{I}^t, \hat{\mathcal{I}^t}),
\end{equation}
where the perceptual loss~\cite{johnson2016perceptual} encourages the preservation of high-level features, and $\lambda$ is a balancing hyperparameter.

\begin{table*}[t]
  \centering
  \resizebox{\textwidth}{!}{%
  \begin{tabular}{lcccccccccccc}
  \hline
   &
    \multicolumn{3}{c}{\textbf{Anisotropic Pixel}} &
    \multicolumn{3}{c}{\textbf{World Scale}} &
    \multicolumn{3}{c}{\textbf{FOV}} &
    \multicolumn{3}{c}{\textbf{Roll}} \\
  \multirow{-2}{*}{\textbf{Model}} &
    SSIM $\uparrow$ &
    LPIPS $\downarrow$ &
    PSNR (M) $\uparrow$ &
    SSIM $\uparrow$ &
    LPIPS $\downarrow$ &
    PSNR (M) $\uparrow$ &
    SSIM $\uparrow$ &
    LPIPS $\downarrow$ &
    PSNR (M) $\uparrow$ &
    SSIM $\uparrow$ &
    LPIPS $\downarrow$ &
    PSNR (M) $\uparrow$ \\ \hline
  AnySplat~\cite{jiang_anysplatfeedforward_2025} &
    0.634 &
    0.376 &
    10.48 &
    0.616 &
    0.346 &
    18.06 &
    0.808 &
    0.163 &
    15.37 &
    0.536 &
    0.446 &
    14.14 \\
  WorldMirror~\cite{liu_worldmirroruniversal_2025} &
    0.578 &
    0.380 &
    11.71 &
    \cellcolor[HTML]{ecf6e5}0.742 &
    \cellcolor[HTML]{ecf6e5}0.250 &
    \cellcolor[HTML]{ecf6e5}22.07 &
    0.855 &
    0.122 &
    18.30 &
    0.552 &
    \cellcolor[HTML]{d8eacb}0.379 &
    16.50 \\
  LVSM~\cite{jin_lvsmlarge_2025} &
    \cellcolor[HTML]{ecf6e5}0.725 &
    \cellcolor[HTML]{ecf6e5}0.235 &
    19.58 &
    0.318 &
    0.633 &
    14.56 &
    0.813 &
    0.119 &
    18.67 &
    \cellcolor[HTML]{ecf6e5}0.588 &
    0.454 &
    \cellcolor[HTML]{ecf6e5}19.54 \\
  RayZer*~\cite{jiang_rayzerselfsupervised_2025} &
    0.578 &
    0.380 &
    11.71 &
    0.428 &
    0.483 &
    14.06 &
    0.730 &
    0.223 &
    13.52 &
    0.337 &
    0.631 &
    10.34 \\
  Ours &
    \cellcolor[HTML]{d8eacb}0.763 &
    \cellcolor[HTML]{d8eacb}0.215 &
    \cellcolor[HTML]{ecf6e5}19.66 &
    \cellcolor[HTML]{d8eacb}0.812 &
    \cellcolor[HTML]{d8eacb}0.169 &
    \cellcolor[HTML]{d8eacb}25.43 &
    \cellcolor[HTML]{d8eacb}0.877 &
    \cellcolor[HTML]{d8eacb}0.104 &
    \cellcolor[HTML]{ecf6e5}20.88 &
    \cellcolor[HTML]{d8eacb}0.629 &
    \cellcolor[HTML]{ecf6e5}0.411 &
    17.53 \\ \hline
  LVSM + aug. &
    0.721 &
    0.252 &
    \cellcolor[HTML]{d8eacb}20.09 &
    0.344 &
    0.666 &
    13.57 &
    \cellcolor[HTML]{ecf6e5}0.870 &
    \cellcolor[HTML]{d8eacb}0.104 &
    \cellcolor[HTML]{d8eacb}21.26 &
    \cellcolor[HTML]{ecf6e5}0.588 &
    0.442 &
    \cellcolor[HTML]{d8eacb}19.79 \\
  Ours + aug. &
    \cellcolor[HTML]{c6e6c0}\textbf{0.784} &
    \cellcolor[HTML]{c6e6c0}\textbf{0.191} &
    \cellcolor[HTML]{c6e6c0}\textbf{20.33} &
    \cellcolor[HTML]{c6e6c0}\textbf{0.823} &
    \cellcolor[HTML]{c6e6c0}\textbf{0.155} &
    \cellcolor[HTML]{c6e6c0}\textbf{25.78} &
    \cellcolor[HTML]{c6e6c0}\textbf{0.890} &
    \cellcolor[HTML]{c6e6c0}\textbf{0.090} &
    \cellcolor[HTML]{c6e6c0}\textbf{21.55} &
    \cellcolor[HTML]{c6e6c0}\textbf{0.702} &
    \cellcolor[HTML]{c6e6c0}\textbf{0.303} &
    \cellcolor[HTML]{c6e6c0}\textbf{20.04} \\ \hline
  \end{tabular}%
  }
\caption{\textbf{Quantitative results on our proposed Consistency Benchmark.} We evaluate model robustness against four types of camera transformations. Our method produces more consistency results with the projective conditioning compared to ray-based~\cite{jin_lvsmlarge_2025,jiang_rayzerselfsupervised_2025} view synthesis models and 3D Gaussian baselines~\cite{jiang_anysplatfeedforward_2025,liu_worldmirroruniversal_2025}. *We use the 24 view checkpoint from RayZer~\cite{jiang_rayzerselfsupervised_2025}, see \cref{sec:benchmark} for details. ``+ aug.'' denotes models fine-tuned with additional camera augmentations for 500 extra steps.}
\label{tab:consistency-benchmark}
\vspace{-4mm}
\end{table*}

\begin{table*}[t]
  \centering
  \resizebox{\textwidth}{!}{%
  \begin{tabular}{lllllllllllll}
  \hline
  \multicolumn{1}{c}{} &
    \multicolumn{3}{c}{\textbf{Small   Overlap}} &
    \multicolumn{3}{c}{\textbf{Medium   Overlap}} &
    \multicolumn{3}{c}{\textbf{Large   Overlap}} &
    \multicolumn{3}{c}{\textbf{Total}} \\
  \multicolumn{1}{c}{\multirow{-2}{*}{\textbf{Model}}} &
    PSNR $\uparrow$ &
    SSIM $\uparrow$ &
    LPIPS $\downarrow$ &
    PSNR $\uparrow$ &
    SSIM $\uparrow$ &
    LPIPS $\downarrow$ &
    PSNR $\uparrow$ &
    SSIM $\uparrow$ &
    LPIPS $\downarrow$ &
    PSNR $\uparrow$ &
    SSIM $\uparrow$ &
    LPIPS $\downarrow$ \\ \hline
  PixelNeRF~\cite{yu_pixelnerfneural_2021} &
    19.27 &
    0.536 &
    0.568 &
    20.38 &
    0.559 &
    0.540 &
    20.94 &
    0.581 &
    0.517 &
    20.26 &
    0.560 &
    0.540 \\
  PixelSplat~\cite{charatan_pixelsplat3d_2024} &
    21.22 &
    0.752 &
    0.225 &
    23.61 &
    0.821 &
    0.162 &
    26.18 &
    0.879 &
    0.115 &
    23.76 &
    0.821 &
    0.164 \\
  MVSplat~\cite{chen_mvsplatefficient_2024} &
    20.67 &
    0.730 &
    0.238 &
    23.97 &
    0.819 &
    0.165 &
    27.32 &
    \cellcolor[HTML]{C6E6C0}\textbf{0.889} &
    \cellcolor[HTML]{ecf6e5}0.112 &
    24.12 &
    0.817 &
    0.168 \\
  NoPoSplat~\cite{ye_nopose_2024} &
    21.58 &
    0.750 &
    0.231 &
    23.67 &
    0.808 &
    0.177 &
    25.84 &
    0.854 &
    0.133 &
    23.78 &
    0.807 &
    0.178 \\
  AnySplat~\cite{jiang_anysplatfeedforward_2025} &
    15.07 &
    0.581 &
    0.411 &
    17.08 &
    0.613 &
    0.350 &
    19.58 &
    0.654 &
    0.281 &
    17.30 &
    0.617 &
    0.345 \\
  Hunyuan-WorldMirror~\cite{liu_worldmirroruniversal_2025} &
    19.03 &
    0.677 &
    0.313 &
    21.40 &
    0.741 &
    0.250 &
    23.95 &
    0.796 &
    0.196 &
    21.55 &
    0.741 &
    0.250 \\
  LVSM (12   layers)~\cite{jin_lvsmlarge_2025} &
    21.58 &
    0.721 &
    0.251 &
    24.49 &
    0.796 &
    0.180 &
    27.38 &
    0.858 &
    0.127 &
    24.60 &
    0.795 &
    0.182 \\
  RayZer*~\cite{jiang_rayzerselfsupervised_2025} &
    20.70 &
    0.669 &
    0.278 &
    22.82 &
    0.737 &
    0.222 &
    24.75 &
    0.789 &
    0.184 &
    22.85 &
    0.735 &
    0.225 \\
  Ours (12   layers) &
    \cellcolor[HTML]{d8eaca}23.64 &
    \cellcolor[HTML]{d8eaca}0.789 &
    \cellcolor[HTML]{d8eaca}0.188 &
    \cellcolor[HTML]{d8eaca}25.60 &
    \cellcolor[HTML]{d8eaca}0.833 &
    \cellcolor[HTML]{d8eaca}0.147 &
    \cellcolor[HTML]{ecf6e5}27.43 &
    0.867 &
    0.116 &
    \cellcolor[HTML]{ecf6e5}25.64 &
    \cellcolor[HTML]{d8eaca}0.832 &
    \cellcolor[HTML]{d8eaca}0.148 \\ \hline
  LVSM (24   layers)~\cite{jin_lvsmlarge_2025} &
    \cellcolor[HTML]{ecf6e5}22.71 &
    \cellcolor[HTML]{ecf6e5}0.765 &
    \cellcolor[HTML]{ecf6e5}0.202 &
    \cellcolor[HTML]{d8eaca}25.60 &
    \cellcolor[HTML]{ecf6e5}0.830 &
    \cellcolor[HTML]{ecf6e5}0.149 &
    \cellcolor[HTML]{d8eaca}28.58 &
    \cellcolor[HTML]{ecf6e5}0.887 &
    \cellcolor[HTML]{d8eaca}0.108 &
    \cellcolor[HTML]{d8eaca}25.74 &
    \cellcolor[HTML]{ecf6e5}0.830 &
    \cellcolor[HTML]{ecf6e5}0.150 \\
  Ours (24   layers) &
    \cellcolor[HTML]{C6E6C0}\textbf{24.98} &
    \cellcolor[HTML]{C6E6C0}\textbf{0.807} &
    \cellcolor[HTML]{C6E6C0}\textbf{0.171} &
    \cellcolor[HTML]{C6E6C0}\textbf{26.88} &
    \cellcolor[HTML]{C6E6C0}\textbf{0.852} &
    \cellcolor[HTML]{C6E6C0}\textbf{0.132} &
    \cellcolor[HTML]{C6E6C0}\textbf{28.60} &
    \cellcolor[HTML]{C6E6C0}\textbf{0.889} &
    \cellcolor[HTML]{C6E6C0}\textbf{0.101} &
    \cellcolor[HTML]{C6E6C0}\textbf{26.90} &
    \cellcolor[HTML]{C6E6C0}\textbf{0.851} &
    \cellcolor[HTML]{C6E6C0}\textbf{0.133} \\ \hline
  \end{tabular}%
  }
  \caption{\textbf{Quantitative evaluation results on the RealEstate10K~\cite{re10k} dataset.} We follow the benchmark splits from NoPoSplat~\cite{ye_nopose_2024}. *We use the 24 view checkpoint from RayZer~\cite{jiang_rayzerselfsupervised_2025}, see \cref{sec:benchmark} for details.}
  \label{tab:quant-re10k}
  \vspace{-5mm}
  \end{table*}

\section{Experiments}
\label{sec:experiments}
\paragraph{Implementation Details.} We follow most architectural design choices outlined in LVSM~\cite{jin_lvsmlarge_2025}. Specifically, our model is a vision transformer with 12 or 24 multi-head self-attention layers. We use a patch size of $p=8$ and latent dimension $d_\text{model}=768$ with 64 dimensions per head. Our model is pretrained for 100,000 steps with an AdamW~\cite{adamw} optimizer. The learning rate is cosine scheduled with a peak value = $10^{-3}$ and 3000 warm-up steps at the beginning. During the fine-tuning stage, the model continues from the pretrained state and utilizes a new warm-up cosine learning rate schedule, with a peak value of $4\times10^{-4}$.

\paragraph{Compute.} Our 12-layer model is pretrained and fine-tuned on 4 NVIDIA 3090 GPUs with a batch size of 6 per GPU for approximately 120 hours. In contrast, the 24-layer model is trained on NVIDIA H100 GPUs for around 1560 GPU-hours, which is  $\sim$7 times lower than the training time of our baseline model~\cite{jin_lvsmlarge_2025}.

\subsection{Consistency Benchmark}
\label{sec:benchmark}

\paragraph{Benchmark Construction.}
To rigorously evaluate the 3D consistency and robustness of view synthesis models under shifts in camera parameter distributions, we construct a new benchmark based on the NoPoSplat~\cite{ye_nopose_2024} benchmark. For each sequence, we apply four types of out-of-distribution transformations to the target cameras while keeping the context views unchanged, simulating real-world usage:
\begin{itemize}
\item \textbf{Anisotropic Pixel:} Modify intrinsics to change the pixel aspect ratio within [0.1, 10].
\item \textbf{World Scale:} Scale the world coordinate system by a constant factor, adjusting camera positions and depth accordingly while keeping ground-truth images unchanged.
\item \textbf{FOV:} Simulate zooming by resizing the target image and updating the focal length in the intrinsics.
\item \textbf{Roll:} Apply an in-plane roll rotation to the target camera.
\end{itemize}
Because these transformations may introduce empty regions in the target frame, PSNR is computed only over valid pixels (``PSNR (M)''), while SSIM and LPIPS~\cite{zhang_unreasonableeffectiveness_2018} are evaluated on the full image.

\paragraph{Comparisons.}
As shown in \cref{tab:consistency-benchmark}, our method consistently outperforms all baselines across all transformations. The largest gain appears in the \textit{World Scale} test, where our model reaches 25.43 PSNR versus LVSM's 14.56, underscoring the brittleness of Pl\"ucker ray conditioning to scale changes. Our method also excels in the \textit{Anisotropic Pixel} and \textit{FOV} settings, achieving higher SSIM and lower LPIPS, reflecting better structural and perceptual fidelity. With an additional 500 steps of camera augmentations (``+ aug.''), our model adapts quickly, whereas ``LVSM + aug.'' shows only limited improvement in short time since these simple transformations can cause significant distribution shift in the Pl\"ucker ray space.

Qualitative comparisons (\cref{fig:qualitative-benchmark}) reinforce these findings. LVSM produces pronounced jagged or grid-like artifacts under roll, FOV, and aspect-ratio variations, and collapses entirely under large world-scale changes. Gaussian-based methods such as AnySplat often generate distortions and holes, likely from inaccurate intrinsics or geometry estimation, while RayZer frequently fails to render the specified viewpoint due to limitations in its Camera Estimator. In contrast, our approach remains stable and geometrically coherent across all transformations, demonstrating the robustness of projective conditioning.

\paragraph{Evaluating RayZer.} RayZer~\cite{jiang_rayzerselfsupervised_2025} only open-sourced the checkpoint trained for the setting with 16 context views and 8 target views ($\mathtt{realestate52K.pt}$). Directly using the provided checkpoint in our evaluation setting leads to degraded results because their model hardcodes the view index into the positional embedding. To ensure a fair comparison, we therefore appended repeated views to the token list with fabricated view indices.

\begin{figure*}[t]
    \centering
    \includegraphics[width=0.9\textwidth]{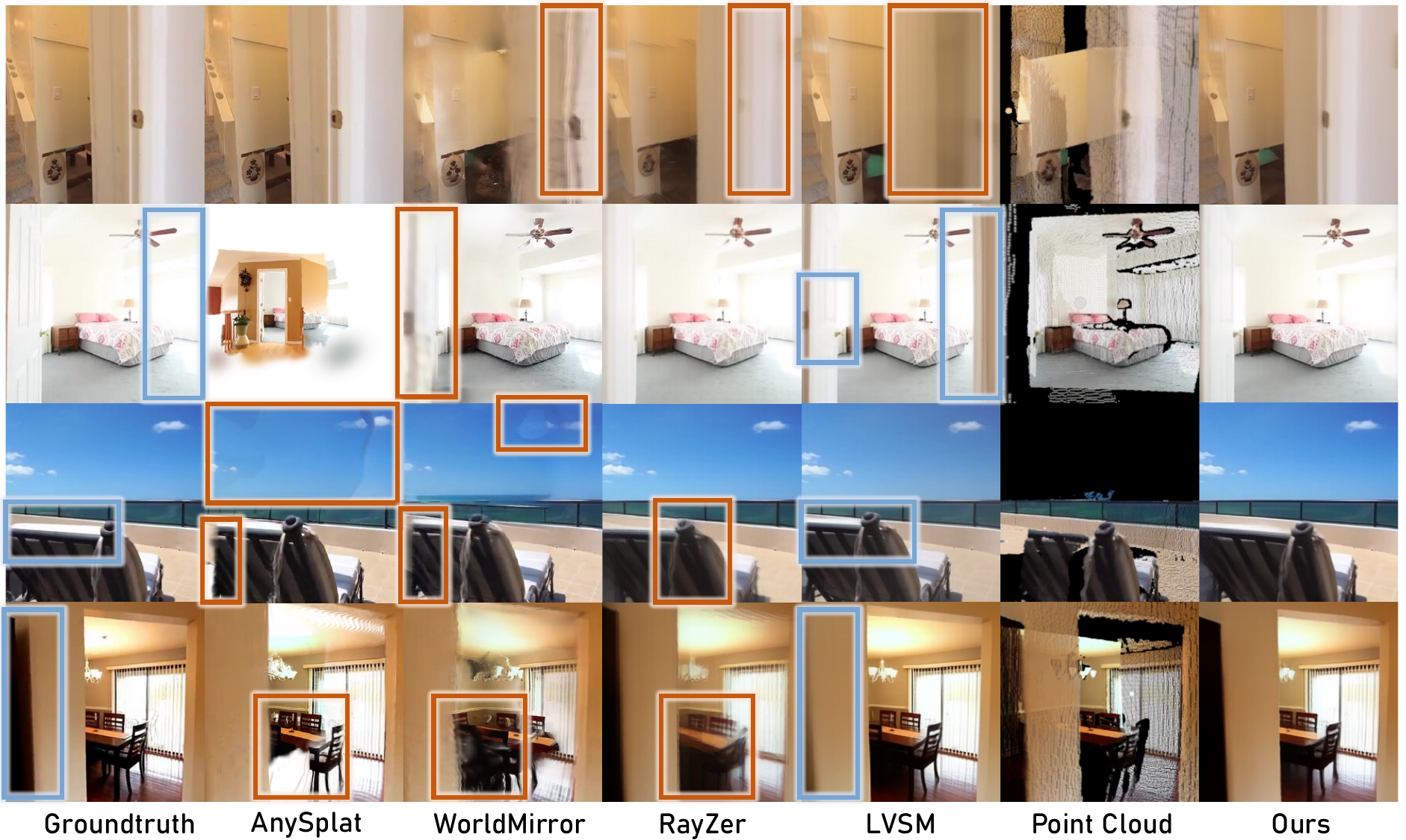}
    \caption{\textbf{Qualitative comparisons on the RealEstate10K~\cite{re10k} dataset.} Our model consistently generates more plausible and geometrically consistent results. We highlight two common failure modes where \textcolor[HTML]{c35f21}{Orange boxes} indicate rendering artifacts like blurriness and ghosting, while \textcolor[HTML]{6e9cee}{blue boxes} suggest geometric errors where models fail to preserve scene structure or render from the correct viewpoint.}
    \label{fig:qualitative}
    \vspace{-4mm}
\end{figure*}

\begin{table}[th]
\centering
\resizebox{\columnwidth}{!}{%
\begin{tabular}{lccc}
\hline
                                                         & \textit{Seen} PSNR      & \textit{Unseen} PSNR    & Full           \\ \hline
AnySplat~\cite{jiang_anysplatfeedforward_2025}           & 16.22          & 11.86          & 15.07          \\
Hunyuan-WorldMirror~\cite{liu_worldmirroruniversal_2025} & 19.65          & 16.91          & 19.03          \\
LVSM~\cite{jin_lvsmlarge_2025}                           & 22.14          & 19.39          & 21.58          \\
Ours                                                     & \textbf{24.30} & \textbf{21.29} & \textbf{23.64} \\
Point Cloud   Image (reference)                          & 15.81          & N/A            & 12.54          \\ \hline
\end{tabular}%
}
\caption{\textbf{Seen vs.\ unseen analysis on RealEstate10K-Small.} We split the target view into \textit{seen} and \textit{unseen} regions based on the projected point cloud image as a visibility mask. Our method captures the view dependent effects in \textit{seen} regions and hallucinates more plausible content in \textit{unseen} regions.}
\label{tab:seen-unseen}
\vspace{-5mm}
\end{table}

\begin{table*}[t]
\centering
\resizebox{\textwidth}{!}{%
\begin{tabular}{cccccclllc}
\hline
\# &
  Model &
  Pretrained on &
  for (steps) &
  Finetune on &
  for (steps) &
  \multicolumn{1}{c}{PSNR (Large)} &
  \multicolumn{1}{c}{PSNR (Medium)} &
  \multicolumn{1}{c}{PSNR (Small)} &
  Total \\ \hline
1 & LVSM & \multicolumn{2}{c}{None} & RealEstate10K & 100K & 27.38 & 24.49 & 21.58 & 24.60 \\
2 & Ours & \multicolumn{2}{c}{None} & RealEstate10K & 100K & 26.96 & 25.11 & 23.06 & 25.13 \\\hline
3 & Ours & RealEstate10K   & 75K    & RealEstate10K & 25K  & 26.28 & 24.78 & 23.02 & 24.78 \\
4 & Ours & DL3DV           & 100K   & RealEstate10K & 200  & 23.36 & 22.37 & 21.03 & 22.32 \\
5 &
  Ours &
  DL3DV &
  100K &
  RealEstate10K &
  \textbf{50K} &
  \textbf{27.43} &
  \textbf{25.60} &
  \textbf{23.64} &
  \textbf{25.64} \\ \hline
\end{tabular}%
}
\caption{\textbf{Ablation studies} on pretraining with different pretraining datasets and finetuning steps. Pretraining on large-scale dataset~\cite{ling2024dl3dv} provides a powerful and generalizable prior for the target domain. Evaluated on the NoPoSplat~\cite{ye_nopose_2024} Large / Medium / Small benchmark.}
\label{tab:pretrain}
\vspace{-5mm}
\end{table*}

\subsection{Sparse View Novel View Synthesis}
\label{sec:quant}
We compare the quality of sparse view novel view synthesis against several baselines, including: 1) NeRF-based method: pixelNeRF~\cite{yu_pixelnerfneural_2021}; 2) Feed-forward 3D Gaussians: PixelSplat~\cite{charatan_pixelsplat3d_2024}, MVSplat~\cite{chen_mvsplatefficient_2024}, NoPoSplat~\cite{ye_nopose_2024}, AnySplat~\cite{jiang_anysplatfeedforward_2025}, and Hunyuan-WorldMirror~\cite{liu_worldmirroruniversal_2025}; 3) View synthesis models: LVSM~\cite{jin_lvsmlarge_2025} and RayZer~\cite{jiang_rayzerselfsupervised_2025}.
As shown in \cref{tab:quant-re10k}, our model consistently surpasses all the baseline methods. Against the strongest 12-layer LVSM, our 12-layer variant delivers small improvements at large overlaps and increasingly larger gains as overlap decreases. Scaling to 24 layers further boosts performance. Compared to the full 24-layer LVSM, our model matches the performances at large overlaps and clearly outperforms it at medium and small overlap setting, indicating stronger robustness to viewpoint variation and (dis)occlusions.

Qualitatively, as illustrated in \cref{fig:qualitative}, our method produces more plausible and geometrically consistent results. Gaussian-based methods, such as AnySplat and WorldMirror, exhibit noticeable rendering artifacts, including blurriness and ghosting, particularly when synthesizing views with large viewpoint changes. Other view synthesis models, like RayZer and LVSM, struggle with geometric consistency and view controllability, often resulting in distorted scenes or failing to render the correct structures.

\paragraph{Seen vs. Unseen Breakdown.}
\cref{tab:seen-unseen} refines the Small-overlap split by separating \emph{seen} and \emph{unseen} target pixels. We classify a target pixel as \emph{seen} if it is covered by the projected point cloud image; 
the remaining pixels are classified as \emph{unseen}. Our method achieves 24.30dB PSNR on \textit{seen} regions and 21.29dB PSNR on \textit{unseen} regions, exceeding LVSM by +2.16dB and +1.90dB PSNR, respectively. This analysis demonstrates that our approach generalizes beyond mere warping, maintaining the largest margin on \emph{unseen} pixels where synthesis must hallucinate newly revealed content. For reference, directly rendering the projected point-cloud image is significantly inferior, underscoring that the improvements stem from learned synthesis rather than simply copying geometry.

\begin{table}[h]
\centering
\resizebox{\columnwidth}{!}{%
\begin{tabular}{cccccc}
\hline
 Time (ms) &
  WorldMirror &
  LVSM &
  RayZer* &
  Ours-12 &
  Ours-24 \\ \hline
Processing &
  74 &
  2.8 &
  18 &
  1.1 &
  2.5 \\
Rendering &
  26 &
  52 &
  56 &
  28 &
  56 \\ \hline
\end{tabular}%
}
\caption{\textbf{Runtime Analysis}. Measured on NVIDIA 3090 GPUs.}
\label{tab:runtime}
\vspace{-2mm}
\end{table}

\begin{figure}[t]
    \centering
    \includegraphics[width=1\linewidth]{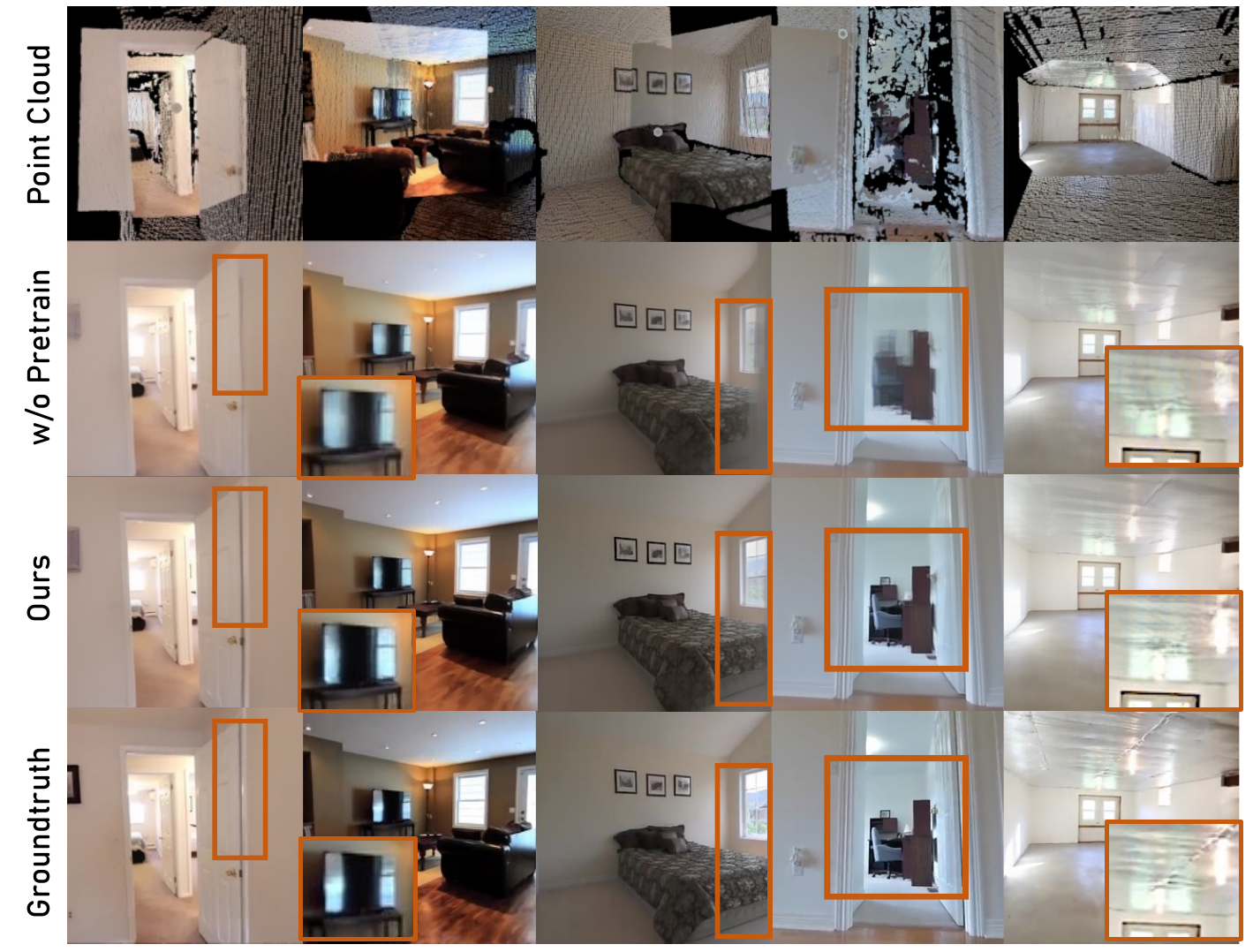}
    \caption{\textbf{Ablation studies} on pretraining. Rendering artifacts are highlighted or zoomed in with \textcolor[HTML]{c35f21}{orange} boxes.}
    \label{fig:ablation-pretrain}
    \vspace{-5mm}
\end{figure}
\paragraph{Runtime Analysis.}
We compare the runtime of our model against several baselines in \cref{tab:runtime}. The analysis is divided into two phases: `Processing  Time', which measures the duration required to construct the 3D scene representation from input views, and `Rendering Time', which measures the time taken to render a single novel view. Our 12-layer model runs in real-time with a significantly lower processing time than the 3D Gaussian-based baseline~\cite{liu_worldmirroruniversal_2025} while our full 24-layer model matches rendering speed of our baseline view synthesis models~\cite{jin_lvsmlarge_2025,jiang_rayzerselfsupervised_2025}.

\subsection{Ablation Studies}
\label{sec:ablations}

We conduct a series of ablation studies to validate the effectiveness of our key design choices.

\paragraph{Pretraining.}
We investigate the impact of pretraining in \cref{tab:pretrain}. We first establish a baseline by training our model from scratch (row 2). Even without pretraining, our model achieves a PSNR of 25.13, already surpassing the baseline method in row 1. To further enhance performance, we explore pretraining with various datasets and different numbers of finetuning steps. While pretraining and finetuning on the same dataset serves as a useful reference, the true breakthrough comes from leveraging a larger, more diverse dataset. By pretraining on the large-scale DL3DV dataset, we provide our model with a powerful and generalizable prior for 3D scene understanding. With just 200 steps of finetuning (row 4), the model rapidly develops foundational rendering capabilities, achieving a PSNR of 22.32. Building upon this strong foundation, a more extensive finetuning schedule of 50K steps (row 5) enables the model to fully adapt its robust prior to the specifics of the target domain.

\paragraph{Other Design Choices.}
\cref{tab:ablation-design} ablates the contributions of several design choices. For DINOv3 prior, we adopt the \texttt{DINOv3-ViT-B} model in our experiments.
Starting with a model that lacks pretraining, DINO prior, and projective conditioning (which corresponds to our baseline~\cite{jin_lvsmlarge_2025}, row1), the addition of projection (row 2) yields consistent improvements (25.20 / 0.811 / 0.177). Incorporating DINO feature priors on top of projection (row 3) further enhances perceptual and structural quality, with only a minor change in PSNR. The combination of pretraining, DINO, and projection (row 4) achieves the best overall results.

\begin{table}[t]
\centering
\resizebox{\linewidth}{!}{%
\begin{tabular}{ccclll}
\hline
Pretrain     & DINO Prior   & Projection & PSNR           & SSIM           & LPIPS          \\ \hline
$\times$     & $\times$     & $\times$         & 24.60          & 0.795          & 0.182          \\
$\times$     & $\times$     & $\checkmark$     & 25.20          & 0.811          & 0.177          \\
$\times$     & $\checkmark$ & $\checkmark$     & 25.13          & 0.816          & 0.163          \\
$\checkmark$ & $\checkmark$ & $\checkmark$     & \textbf{25.64} & \textbf{0.832} & \textbf{0.148} \\ \hline
\end{tabular}%
}
\caption{\textbf{Ablation studies} on pretraining, the use of DINO feature priors, and the projective conditioning. Pretain and evaluation are on the DL3DV~\cite{ling2024dl3dv} and RealEstate10K~\cite{re10k} datasets respectively.}
\label{tab:ablation-design}
\vspace{-5mm}
\end{table}

\section{Conclusion}
\label{sec:conclusion}

This paper investigates the input space representation for feed-forward view synthesis models. The direct encoding of camera parameters using Pl\"ucker ray maps can introduce sensitivity to camera transformations and coordinate system choices, which negatively impacts generalization and 3D consistency. We propose projective conditioning, an approach that models the quotient space of the configuration space, independent of the coordinate system. This reframes view synthesis as a 2D image-to-image mapping, improving camera control and robustness to various transformations. Additionally, we introduce a masked auto-encoding pretraining strategy that leverages the 2D nature of projective conditioning within a self-supervised learning framework, facilitating effective learning from large-scale uncalibrated video data. Experiments conducted on a custom out-of-distribution benchmark—including roll, field of view changes, and scaling—demonstrate state-of-the-art performance in 3D consistency and rendering quality compared to existing baselines. Future work may explore potential extensions to dynamic scenes.

\clearpage
\IfWithSupplTF{\section{Acknowledgment}
We sincerely thank Haozhe Lou, Jiajun Jiang for the suggestions on theoretical analysis; Haoxuan Wang for the assistance and valuable feedbacks on figure designs; Handi Yin, Haonan He, Bonan Liu, Jianteng Chen, Runyi Yang, Liyi Luo, Xuanchao Peng, Yiming Zhu, Shibo Wang and Xinyuan Xu for fruitful discussions and proofreading.
}{}
{
    \small
    \bibliographystyle{ieee_with_arxiv}
    \bibliography{main,wuzirui_bibtex}
}
\IfWithSupplTF{\IfWithSupplTF{\appendix}{
\setcounter{page}{1}
\maketitlesupplementary
}

\section{Additional Experiments}
\paragraph{Ablation Studies on Positional Embedding.}

\cref{tab:ablation-rope} highlights the importance of Rotary Position Embedding (RoPE).
While adding RoPE to the LVSM baseline (`LVSM + RoPE') yields only a slight PSNR improvement, omitting positional encoding in our architecture causes a substantial performance drop. To illustrate this failure mode, \cref{fig:toy-example} presents a toy experiment where we overfit a single-object scene: without RoPE, the model collapses to predicting completely identical patches over the empty regions, regressing to the mean background color of the ground-truth image.

\begin{figure}[h]
    \centering
    \includegraphics[width=0.5\textwidth]{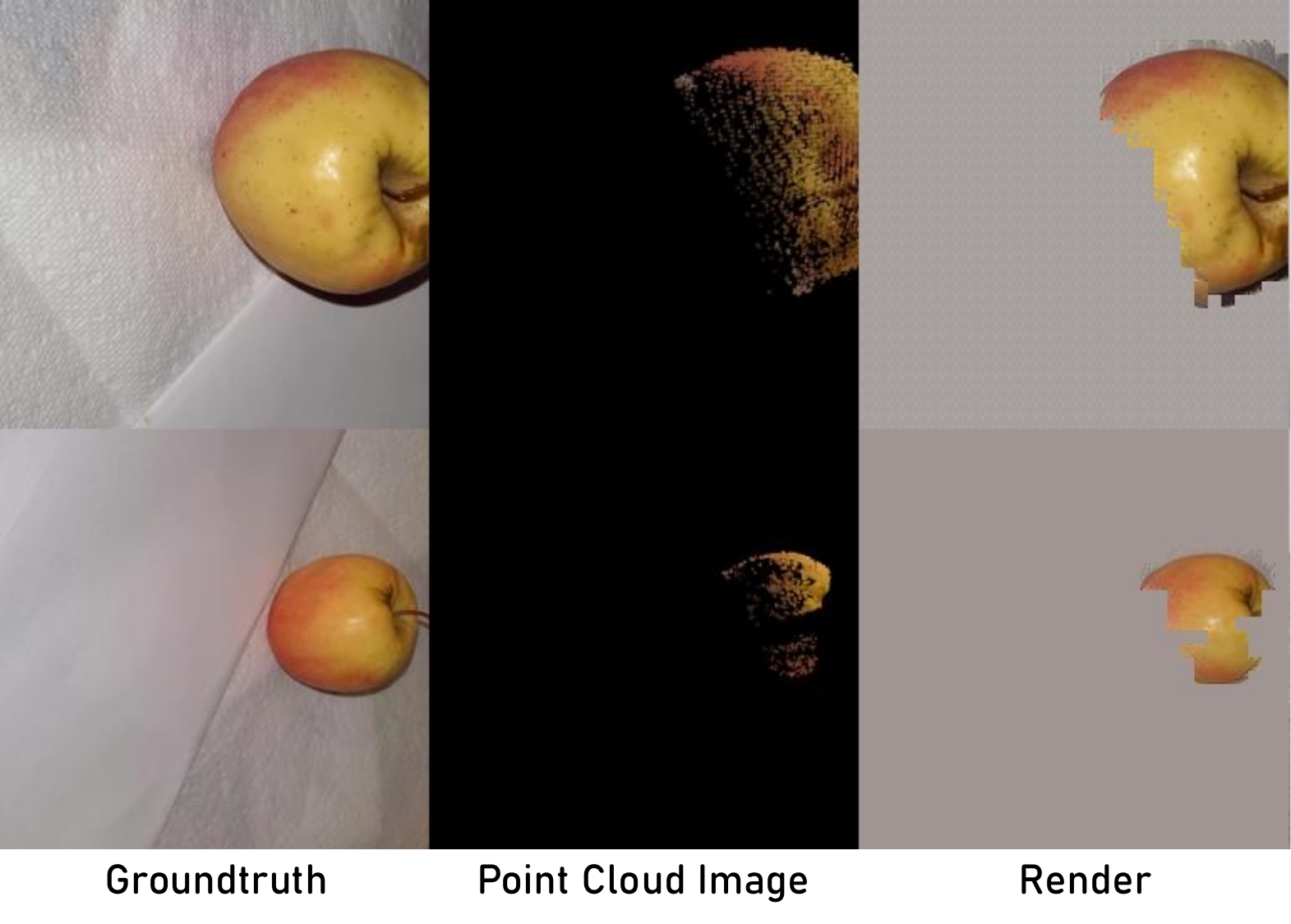}
    \caption{Without RoPE, the model produces degraded results on the identical patches.}
    \label{fig:toy-example}
\end{figure}

\begin{table}[ht]
\centering
\resizebox{\columnwidth}{!}{%
\begin{tabular}{lllll}
\hline
                & LVSM  & LVSM + RoPE & Ours w/o RoPE & Ours  \\ \hline
PSNR $\uparrow$ & 25.39 & 25.88       & 21.18         & 30.03 \\ \hline
\end{tabular}%
}
\caption{\textbf{Ablation studies} on the use of RoPE~\cite{su_roformerenhanced_2023}.}
\label{tab:ablation-rope}
\end{table}

\paragraph{Additional Comparisons with LVSM~\cite{jin_lvsmlarge_2025}.}
We show more qualitative comparisons with LVSM~\cite{jin_lvsmlarge_2025} on the RealEstate10K dataset~\cite{re10k} in \cref{fig:more_comparison}. Without direct geometric cue from the projected point cloud, LVSM often produces wrong prediction on geometry.

\paragraph{Results on the pretraining and the finetuning stage.}
We also show results on the MAE pretraining stage and the finetuning stage in \cref{fig:mae_results} and \cref{fig:finetune_results} respectively.

\begin{figure}[t]
    \centering
    \includegraphics[width=0.5\textwidth]{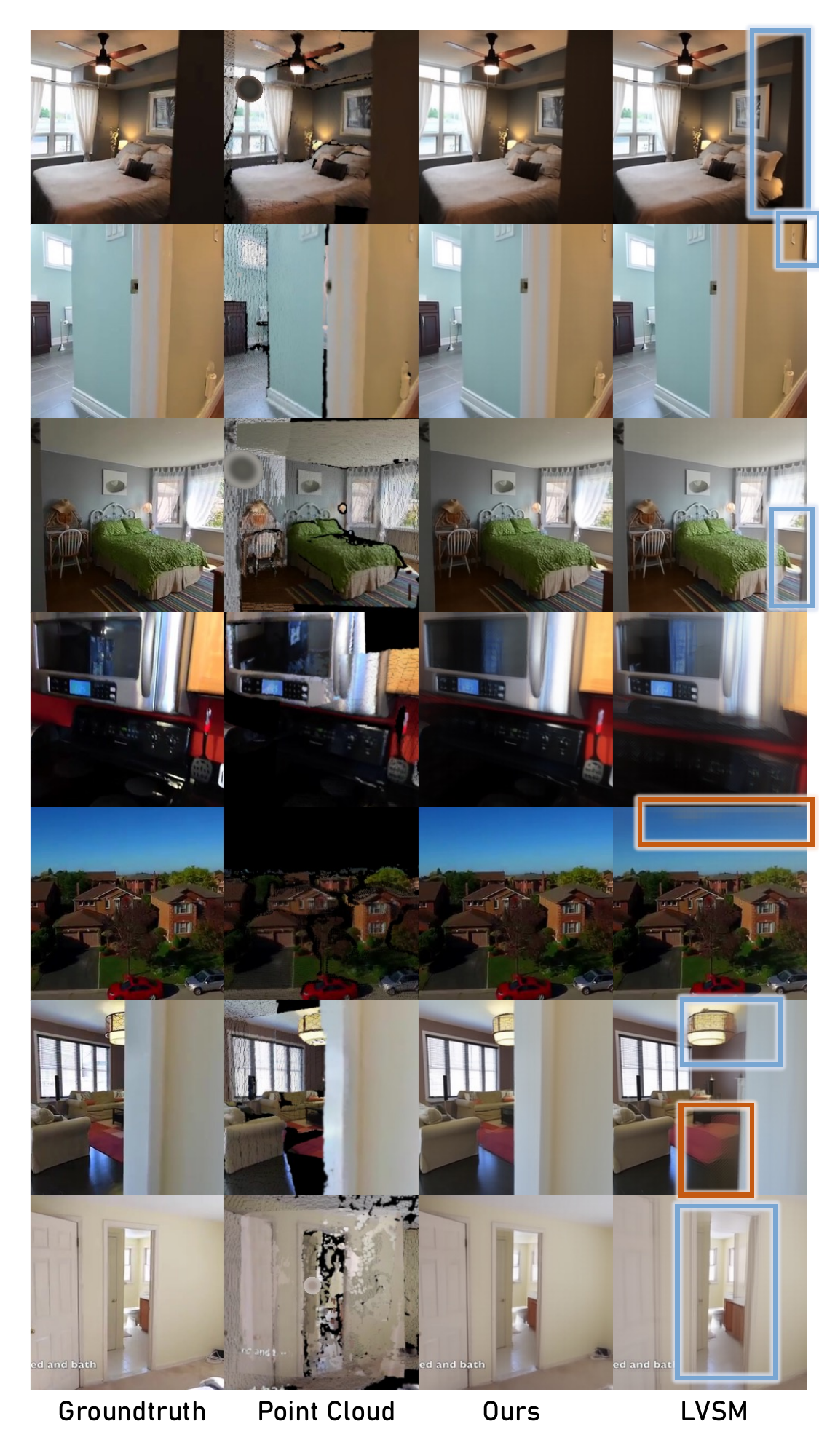}
    \caption{More qualitative comparisons with LVSM~\cite{jin_lvsmlarge_2025} on the RealEstate10K dataset~\cite{re10k}.}
    \label{fig:more_comparison}
\end{figure}

\begin{figure*}[t]
    \centering
    \includegraphics[width=\textwidth]{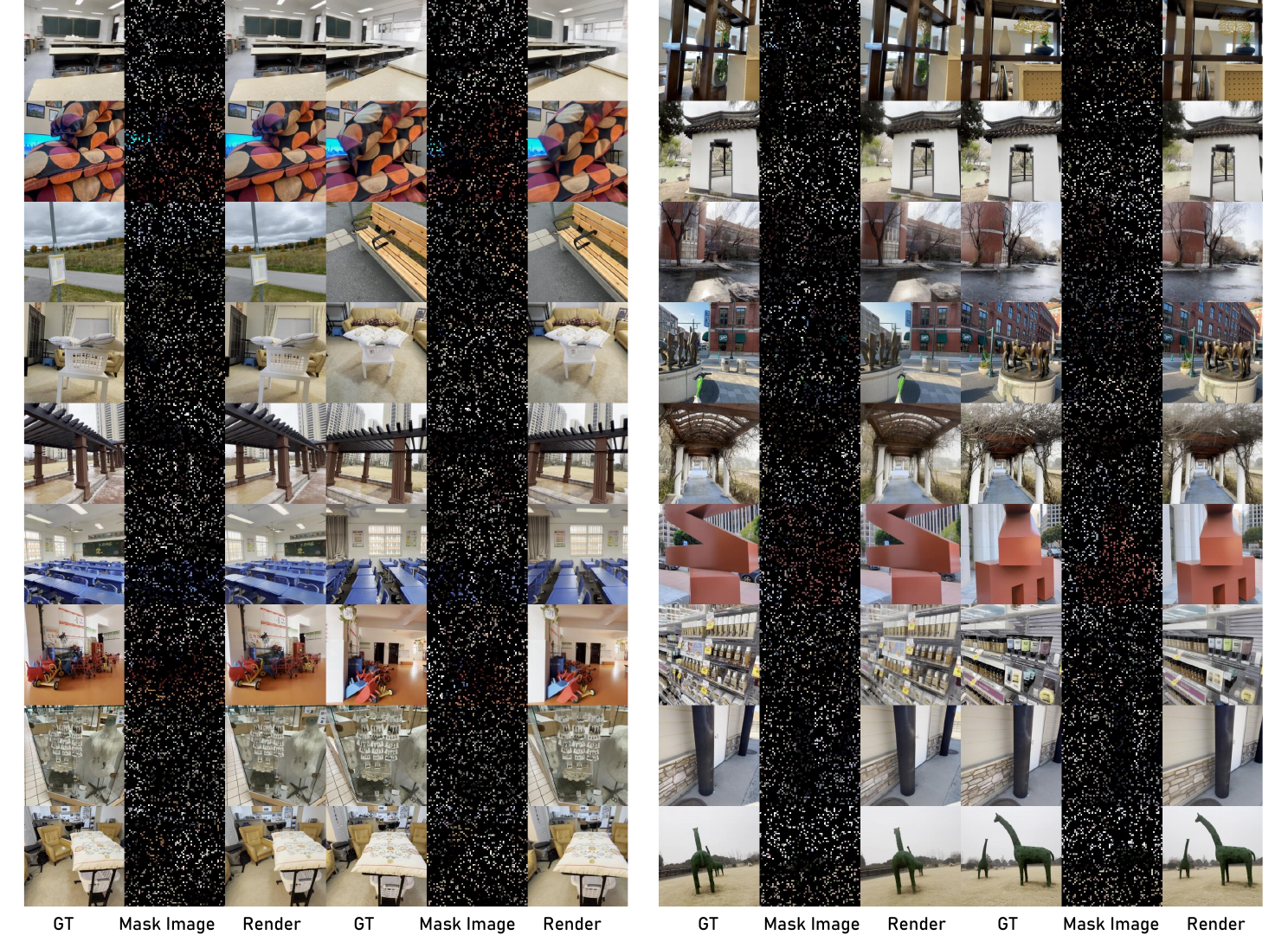}
    \caption{Additional results on the MAE pretraining stage.}
    \label{fig:mae_results}
\end{figure*}

\begin{figure*}[t]
    \centering
    \includegraphics[width=\textwidth]{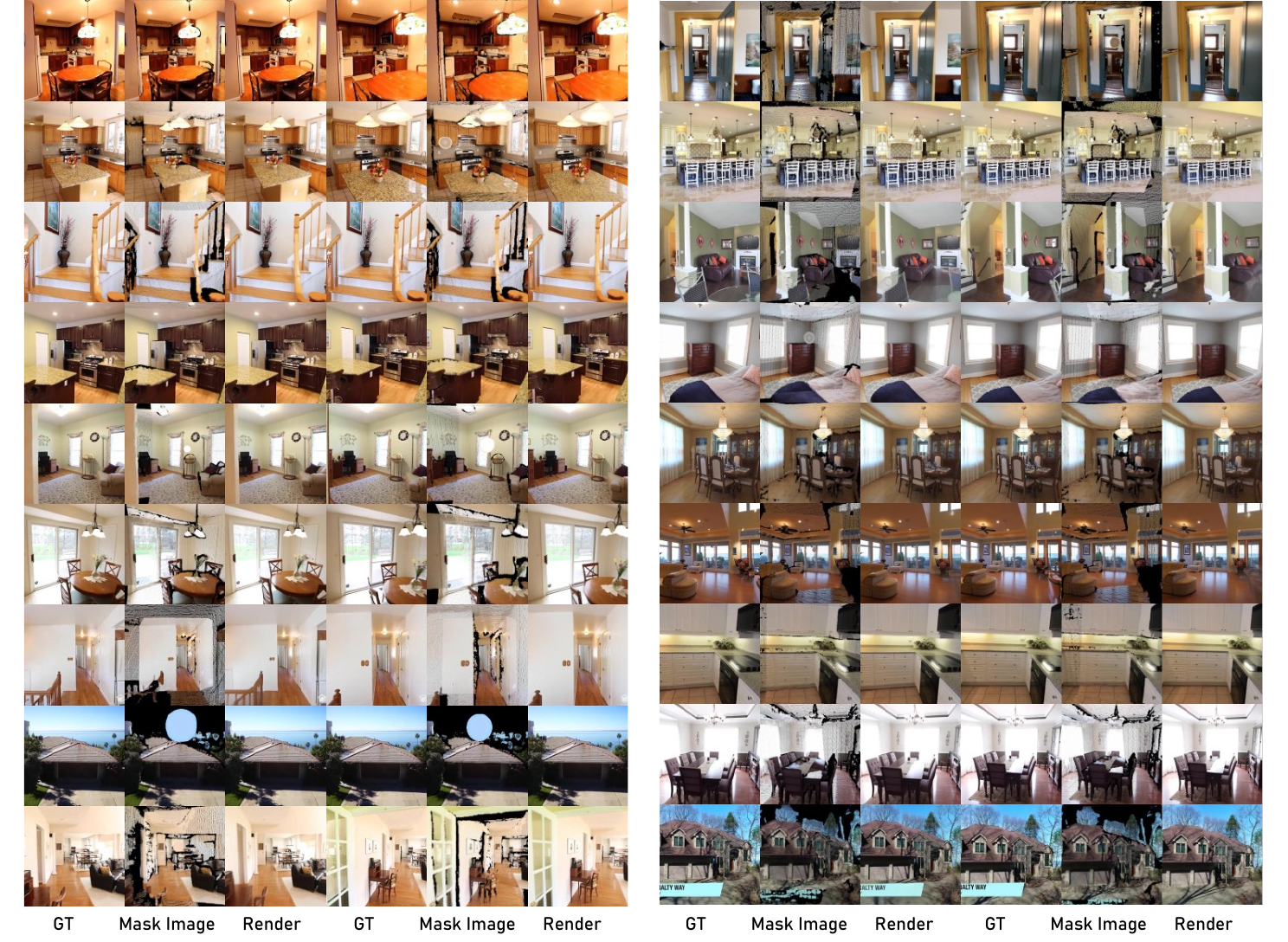}
    \caption{Additional results on the finetuning stage.}
    \label{fig:finetune_results}
\end{figure*}

\paragraph{Object-centric Experiments.}
We further evaluate our method on an object-centric dataset G-Objaverse~\cite{objaverse, qiu2024richdreamer, zuo2024sparse3d}, which contains the rendered G-buffers of the objects in the Objaverse dataset~\cite{objaverse} in around 30 view directions. We follow the same pre-training then finetuning strategy as in the main paper, and the results are shown in \cref{fig:object-centric}.
For object-level scenes, we additionally add a layer of random colored Gaussians behind the depth surface to address ambiguities in the projected point cloud image.

\begin{figure*}[p]
    \centering
    \includegraphics[width=\textwidth]{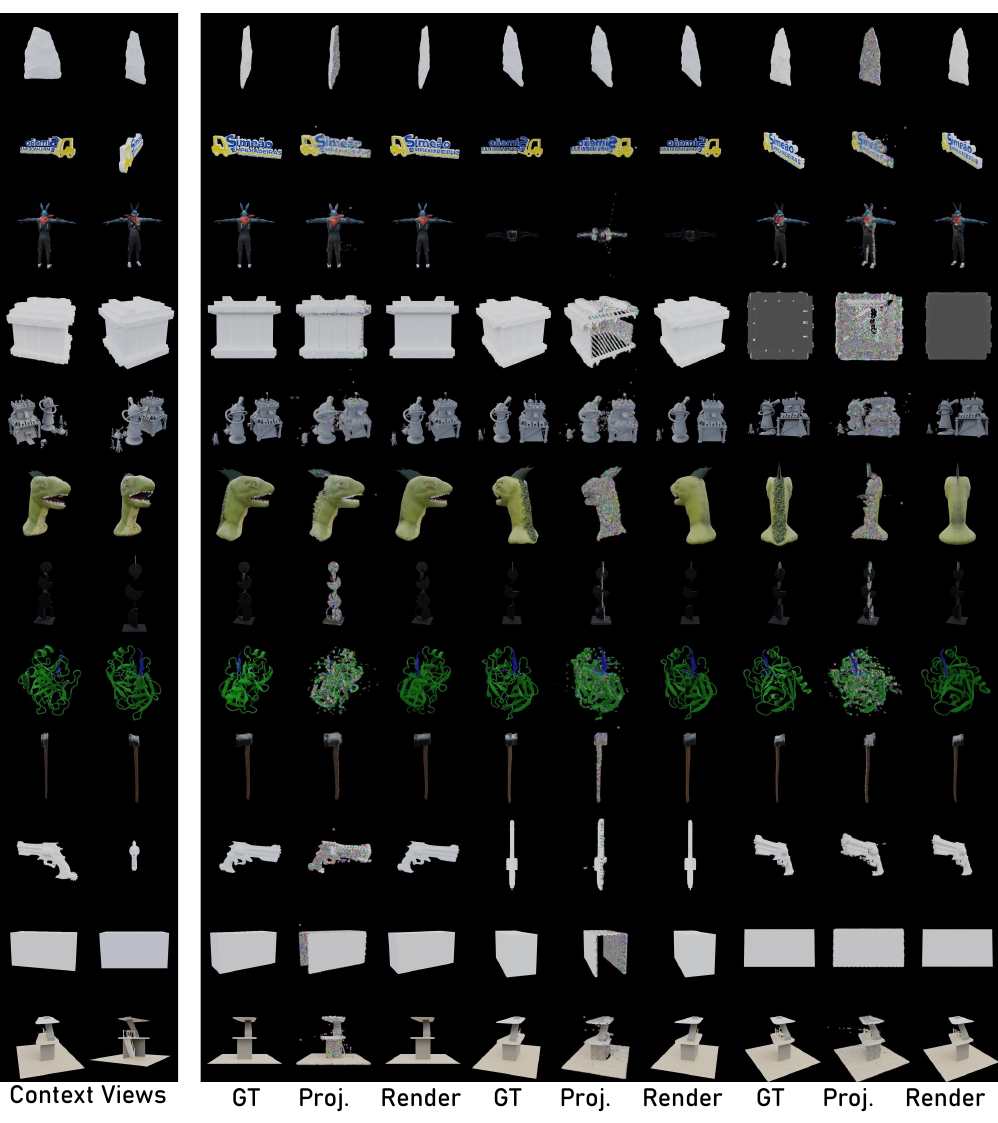}
    \caption{Results on G-Objaverse dataset~\cite{objaverse, qiu2024richdreamer, zuo2024sparse3d}. Different from scene-level datasets, we additionally add a random colored layer behind the seen surfaces to prevent symmetry ambiguities.}
    \label{fig:object-centric}
\end{figure*}

\section{Limitations}
We discuss the limitations of our method in this section.
First, similar to prior regression-based methods~\cite{jin_lvsmlarge_2025,jiang_rayzerselfsupervised_2025}, our method only interpolates between the context views, its ability to hallucinate unseen regions is limited. Although we show better performance on the unseen regions (\IfWithSupplTF{\cref{tab:seen-unseen}}{Tab. 3 in main paper}), it is still restricted to regressing the ``average'' contents across the training dataset (e.g., completing the unseen region of a wall or a floor). Future work could consider combining with generative models~\cite{zheng_diffusiontransformers_2025} to generate more diverse and realistic novel views.

Second, our method is restricted to static scenes, when presented with dynamic objects, the model can produce artifacts like ghosting and blurriness, or inconsistent results across different frames. Though the pretraining stage does not impose static assumption, more diverse training data and fine-tuning strategy are necessary to handle dynamic scenes in our pipeline.

\section{Dolly Zoom Camera Motion}
\label{app:dolly-zoom}

The dolly zoom (also known as the Hitchcock shot~\cite{truffaut1986hitchcock}) is a camera motion where the camera is translated along its viewing
direction while the focal length is adjusted so that a chosen object keeps a
constant image size. This creates the characteristic effect that the foreground
object stays fixed in scale while the background appears to expand or contract.

We model the camera with intrinsics $K = \mathrm{diag}(f_x, f_y, 1)$ and principal point $\mathbf{c} = (c_x,c_y)$.
For a 3D point $\mathbf{X} = (X,Y,Z)^\top$ in camera coordinates, the pinhole
projection is
\begin{equation}
\lambda \begin{bmatrix} u \\ v \\ 1 \end{bmatrix}
= K \begin{bmatrix} X \\ Y \\ Z \end{bmatrix},
\quad
u = f_x \frac{X}{Z} + c_x,\; v = f_y \frac{Y}{Z} + c_y .
\end{equation}
Thus the apparent size of an object at depth $Z$ scales proportionally to
$f_y / Z$.

Let $\mathbf{C}_0$ be the initial camera center and let
$\mathbf{n}_0 \in \mathbb{R}^3$ denote the unit forward direction
(the third column of the rotation matrix $R_0$).
We pick an anchor point $\mathbf{X}_\star$ on the object whose size we wish to
keep fixed. Its initial depth is:
\begin{equation}
Z_0 = \mathbf{n}_0^\top (\mathbf{X}_\star - \mathbf{C}_0).
\end{equation}
During a dolly zoom, the camera is translated along $\mathbf{n}_0$ to:
\begin{equation}
\mathbf{C}(t) = \mathbf{C}_0 + \Delta(t)\,\mathbf{n}_0 ,
\end{equation}
while the orientation is kept constant, $R(t) = R_0$.
The depth of the anchor point in the new camera is then:
\begin{equation}
Z(t) = \mathbf{n}_0^\top (\mathbf{X}_\star - \mathbf{C}(t))
     = Z_0 - \Delta(t).
\end{equation}

To keep the anchor’s image size constant, we require that its scale factor
$f_y(t)/Z(t)$ remains equal to the initial value $f_{y_0}/Z_0$:
\begin{equation}
\frac{f_y(t)}{Z(t)} = \frac{f_{y_0}}{Z_0}
\;\rightarrow\;
f_y(t) = f_{y_0}\,\frac{Z_0 - \Delta(t)}{Z_0}.
\label{eq:dolly-fy}
\end{equation}
Equation~\eqref{eq:dolly-fy} is the core constraint of the dolly zoom: as the
camera moves closer to the object ($\Delta(t) > 0$), the focal length must
decrease to preserve the ratio $f_y/Z$; moving the camera away requires
increasing the focal length.

If we parameterize the camera by its vertical field of view $\theta(t)$ instead of
$f_y(t)$, for an image of height $H$ pixels:
\begin{equation}
f_y(t) = \frac{H}{2 \tan(\theta(t)/2)}.
\end{equation}
Combining this with \eqref{eq:dolly-fy} yields
\begin{equation}
\tan\!\left(\frac{\theta(t)}{2}\right)
=
\tan\!\left(\frac{\theta_0}{2}\right)
\frac{Z_0}{Z_0 - \Delta(t)},
\end{equation}
where $\theta_0$ is the initial field of view.
In practice, we pick an anchor frame, estimate $Z_0$ for a reference pixel
(e.g., the image center), and then, for each target field of view $\theta(t)$,
translate the camera center by $\Delta(t)\,\mathbf{n}_0$ and adjust the
intrinsics according to the relations above. This realizes a physically
consistent dolly zoom trajectory in a standard pinhole camera model.
}{}

\end{document}